\documentclass[letterpaper]{article} 
\usepackage{aaai2026}  
\usepackage{times}  
\usepackage{helvet}  
\usepackage{courier}  
\usepackage[hyphens]{url}  
\usepackage{graphicx} 
\usepackage{natbib}  
\usepackage{caption} 
\usepackage{color}
\usepackage{colortbl}
\usepackage{booktabs} 
\usepackage{xcolor}    
\usepackage{graphicx}  
\usepackage{subcaption}
\usepackage{amssymb}
\usepackage{amsmath}
\usepackage{amsfonts}

\usepackage{algorithm}
\usepackage{algorithmic}

\usepackage{newfloat}
\usepackage{listings}
\DeclareCaptionStyle{ruled}{labelfont=normalfont,labelsep=colon,strut=off} 
\floatstyle{ruled}
\newfloat{listing}{tb}{lst}{}
\floatname{listing}{Listing}
\title{P-SLCR: Unsupervised Point Cloud Semantic Segmentation via Prototypes Structure Learning and Consistent Reasoning}
\author{
	Lixin Zhan\textsuperscript{\rm 1, 2}, 
	Jie Jiang\textsuperscript{\rm 1, 2}\thanks{Corresponding author.}, 
	Tianjian Zhou\textsuperscript{\rm 1, 2}, 
	Yukun Du\textsuperscript{\rm 1}, 
	Yan Zheng\textsuperscript{\rm 1}, 
	Xuehu Duan \textsuperscript{\rm 1, 2}
}
\affiliations{
    \textsuperscript{\rm 1}College of Systems Engineering, National University of Defense Technology\\
    \textsuperscript{\rm 2}Laboratory for Big Data and Decision, National University of Defense Technology\\

    zhanlixin98@outlook.com, jiejiang@nudt.edu.cn, zhoutianjian2000@nudt.edu.cn, duyukun-nudt@outlook.com, zhengyan24@nudt.edu.cn, duanxuehu24@nudt.edu.cn
}

\usepackage{bibentry}

\begin{document}

\maketitle

\begin{abstract}
Current semantic segmentation approaches for point cloud scenes heavily rely on manual labeling, while research on unsupervised semantic segmentation methods specifically for raw point clouds is still in its early stages. Unsupervised point cloud learning poses significant challenges due to the absence of annotation information and the lack of pre-training. The development of effective strategies is crucial in this context. In this paper, we propose a novel prototype library-driven unsupervised point cloud semantic segmentation strategy that utilizes Structure Learning and Consistent Reasoning (P-SLCR). 
First, we propose a Consistent Structure Learning to establish structural feature learning between consistent points and the library of consistent prototypes by selecting high-quality features. 
Second, we propose a Semantic Relation Consistent Reasoning that constructs a prototype inter-relation matrix between consistent and ambiguous prototype libraries separately. This process ensures the preservation of semantic consistency by imposing constraints on consistent and ambiguous prototype libraries through the prototype inter-relation matrix.  
Finally, our method was extensively evaluated on the S3DIS, SemanticKITTI, and Scannet datasets, achieving the best performance compared to unsupervised methods. Specifically, the mIoU of 47.1\% is achieved for Area-5 of the S3DIS  dataset, surpassing the classical fully supervised method PointNet by 2.5\%.
\end{abstract}

\begin{links}
    \link{Code}{https://github.com/lixinzhan98/P-SLCR} 
\end{links}

\section{Introduction}
\label{sec:intro}

Semantic segmentation of point clouds is a crucial task in computer 3D vision, representing fundamental research with growing significance across various applications in recent years. Since the inception of PointNet \cite{qi2017pointnet}, there has been a surge in neural network-based semantic segmentation research for point clouds \cite{2019RandLA, zhao2021point, ZHAN2023103259, wu2024point}, leading to significant enhancements in the accuracy of point segmentation \cite{huang2024opoca, li2023semantic}. However, these methodologies heavily rely on manual annotation, with unstructured 3D data annotation typically requiring greater human and material resources than annotating 2D images.
Certain approaches alleviate this issue by utilizing a limited set of point cloud labels \cite{liu2022less, zhang2021weakly, shi2022weakly, unal2022scribble, hu2022sqn, zhan2025bgc, li2022hybridcr, zhan2025qpcr} during model training. Despite achieving promising outcomes, these methods still depend on manual annotations for data alignment. Extending to new scenarios is challenging due to the need for manual labeling, a highly time-consuming process. While unsupervised methods for 2D images  \cite{ji2019invariant, cho2021picie} are booming, unsupervised understanding of 3D data is just beginning. Some researchers have also achieved unsupervised learning of 3D data through domain adaptive strategies \cite{zhao2021epointda, saltori2023compositional}, but our study does not rely on any transfer learning.
\begin{figure}
	\centering
	\includegraphics[width=0.47\textwidth]{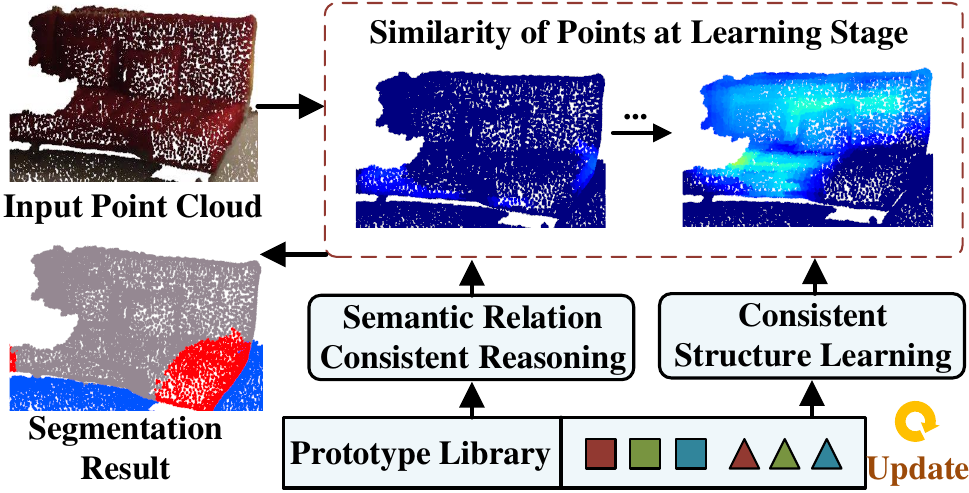} 
	\caption{P-SLCR architecture for structural learning and consistent reasoning through a learnable prototype-based library. The consistent points are continuously expanding and increasingly resembling the prototype.}
	\label{imgtop}
\end{figure}

Our study aims to delineate 3D features without the manual annotation. While numerous methods have emerged to tackle unsupervised point cloud semantic segmentation tasks, representative approaches include PointDC \cite{chen2023pointdc}, GrowSP \cite{zhang2023growsp}, U3DS$^3$ \cite{liu2024u3ds3} and LogoSP \cite{zhang2025logosp}. GrowSP posits that individual 3D points often lack practical significance, while a cluster or set of 3D points holds practical relevance. GrowSP \cite{zhang2023growsp} and U3DS$^3$ \cite{liu2024u3ds3} render the unsupervised task \cite{chen2023pointdc, wu2023masked} of large-scale point cloud analysis feasible through the application of over-segmentation for clustering, employing superpoint segmentation and superpoint region expansion. However, the pseudo-labels generated by the clustering algorithm are not completely trustworthy, and the direct use of the full set of pseudo-labels to supervise network learning may not be conducive to distinguishing salient features across categories. In this way, the prototype features are not representative and do not fully exploit the structural information of the point cloud.

To alleviate the above problems, we propose an unsupervised method for semantic segmentation of point clouds via Prototypes Consistent Structure Learning and Reasoning strategy. As shown in Fig. \ref{imgtop}, we built two prototype libraries, the consistent prototype library, and the ambiguous prototype library. These libraries are updated by the EMA \cite{tarvainen2017mean} algorithm based on the local clustering centers of midpoint batches. Consistent features typically demonstrate higher accuracy than ambiguous features, facilitating the utilization of confidence in selecting high-quality features and establishing a structural similarity learning connection between consistent features and prototypes. The consistent prototype prioritizes learning robust features within each category and can guide the learning of ambiguous prototypes. Semantic features of both consistent and ambiguous prototypes are maintained consistently through constraints on the consistent prototype. The learning process is dynamic,  with the potential for all ambiguous points to be gradually included in the consistent set through continuous learning, thus completing the feature space division and achieving precise semantic segmentation. Our contribution is as follows:   
\begin{itemize}
	\item We propose an innovative unsupervised framework for semantic segmentation of point clouds, emphasizing consistent structure learning and consistent reasoning guided by a dynamic prototype library.
	\item We propose a consistent structure learning that utilizes plausibility to select high-quality features and establish consistent structure learning between consistent point features and a library of consistent prototypes.
	\item We propose a semantic relation consistent reasoning designed to assist in training consistent prototypes, guiding the learning process for ambiguous prototypes. This constraint can be applied to uphold semantic feature consistency within both the consistent and ambiguous prototype libraries.
\end{itemize}

\section{Related Work}

\subsection{Fully Supervised Semantic Segmentation}
Fully supervised point cloud semantic segmentation has achieved great success in recent years. It can be categorized into view-based methods \cite{tatarchenko2018tangent, milioto2019rangenet++, kundu2020virtual}, voxel-based methods \cite{meng2019vv, zhu2021cylindrical}, and point-based methods. The first two categories convert sparse data into regular data, inevitably losing structural information. Point-based methods, stemming from PointNet \cite{qi2017pointnet} and PointNet++ \cite{qi2017pointnet++}, advocate for the employment of permutation invariant functions to extract point cloud features directly. Follow-up work RandLA-Net \cite{2019RandLA} balances efficiency and segmentation accuracy in handling large-scale semantic segmentation of point clouds. The advent of Transformer \cite{vaswani2017attention} has spurred rapid advancements in  Transformer-based approaches \cite{zhao2021point, wu2022point, wu2024point} within point cloud semantic segmentation,  yielding promising outcomes. However, these methods rely on a large amount of annotated information, which is time-consuming and expensive.

\begin{figure*}
	\centering
	\includegraphics[width=0.98\textwidth]{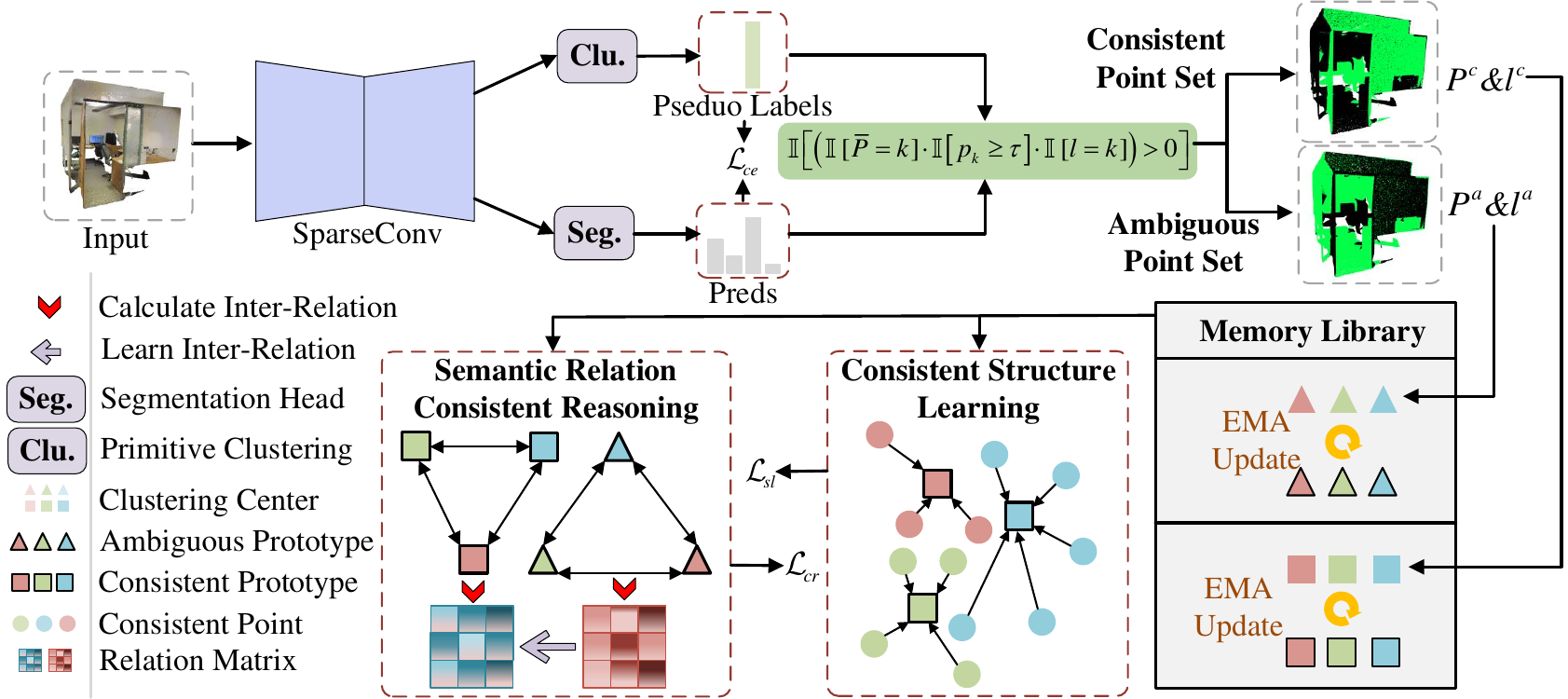} 
	\caption{An overview of our P-SLCR with features extracted by SparseConv, pseudo-labels and predictions obtained by clustering and segmenting the header. The proposed reliability classifies the scene into consistent and ambiguous points and establishes an updatable prototype library to construct consistent structure learning and semantic relation consistent reasoning based on the prototype library. }
	\label{img1}
\end{figure*}

\subsection{Weakly Supervised Semantic Segmentation}
To reduce the burden of complete annotation, researchers have shifted their focus toward training semantic segmentation models with a limited number of annotations. Some of these methods use fewer 3D labels \cite{liu2022less, shi2022weakly, unal2022scribble, hu2022sqn, li2022hybridcr}, and others use a small number of fully labeled scenes \cite{chibane2022box2mask, liu2022weakly, ren20213d}. To enrich annotation information, one may leverage consistency perturbation \cite{zhang2021perturbed} or employ pseudo-label generation \cite{cheng2021sspc}. The consistency perturbation relies on ensuring that perturbed samples maintain consistency with the predictions made for the original sample. The methodology for generating pseudo-labels entails utilizing predicted values from unlabeled points to formulate labels that are integrated into the model for training. Some approaches employ self-supervised pre-training \cite{hou2021exploring, zhang2022self}, and supervised signals can originate from diverse modalities, including 2D images \cite{robert2022learning, shin2022mm} or language models \cite{zhang2022pointclip, rozenberszki2022language}. Nevertheless, these methods still need manual annotation for data alignment, preventing the network model from autonomously discerning semantic classes.

\subsection{Unsupervised Semantic Segmentation}
Several works have proposed methods for 2D unsupervised semantic segmentation \cite{ji2019invariant, cho2021picie}. Nevertheless, the most current methods are inapplicable to point clouds because of the distinctions between images and point clouds. Research has initiated exploration into unsupervised semantic segmentation of 3D point clouds, yielding promising results. GrowSP \cite{zhang2023growsp} represents a pioneering effort in unsupervised semantic segmentation of point clouds. The approach involves initially segmenting superpoints using VCCS \cite{papon2013voxel}, synthesizing superpoint features,  and subsequently clustering the superpoints while assigning pseudo-labels. During the later stages of training, larger superpoints can be synthesized. U3DS$^3$ \cite{liu2024u3ds3} establishes a generalized unsupervised segmentation method by means of spatial clustering, iterative training by clustering cosine distances between the center of mass and points, and expanding the training data by color changes. However, none of them offer effective coaching strategies. Pseudo-labels generated by clustering algorithms are not entirely consistent, and violently using all of them to supervise network learning may be detrimental to distinguishing salient features of a category. The semantic prototype is not universal and does not fully exploit the structural information of the point cloud.


\section{P-SLCR}
\subsection{Overview}
%

Fig. \ref{img1} illustrates the overall framework of our proposed \textbf{P-SLCR} model, which performs unsupervised point cloud semantic segmentation through \textit{prototype library-driven structure learning} and \textit{semantic relation-consistent reasoning}. Given an input point cloud, a feature extractor $G$ encodes geometric and color cues into point-wise embeddings, which are aggregated into semantically homogeneous superpoints. P-SLCR maintains a \textit{dual prototype library} that dynamically separates reliable (consistent) and uncertain (ambiguous) representations according to feature reliability. The consistent prototype library captures stable semantics across scenes, while the ambiguous one serves as a buffer to model uncertain regions and guide their refinement. A structure-consistent reasoning module enforces semantic and relational alignment between the two prototype sets, encouraging globally coherent segmentation. Through iterative optimization, refined pseudo-labels generated from the prototype reasoning loop are fed back to supervise $G$, forming a closed learning cycle that progressively enhances both feature quality and segmentation consistency.

\subsection{Separation of Reliable Points}
Without loss of generality, for points in a point cloud scene, can obtain the features $f$ obtained after the network $G$, the pseudo-labels $\boldsymbol{l}$, and the initial prototypes $\boldsymbol{\mu}$, which are usually the clustering centers of the categories. It is reasonable to assume (Appendix for details.) that if the network $G$ is adequately robust and the prototype $\boldsymbol{\mu}$ is also robust. Theoretically, the local clustering centers predicted by the network $G$ for each batch are expected to be infinitesimally close to the corresponding prototype $\boldsymbol{\mu}$, reflecting local consistency during training. 

In reality, $G$ is not a highly efficient learner, and $\boldsymbol{\mu}$ lacks the required robustness. This approach can be executed as a training strategy for the model.

However, it is not desirable to directly and crudely calculate the clustering center in a single batch based on pseudo-labels. The pseudo-label $\boldsymbol{l}$ assigned by the clustering may lack credibility and contain numerous errors. Therefore, we introduce the concept of consistent points, which mainly filters out points with high confidence. Formally, the consistent points should be satisfied that the clustering assigned pseudo-label $\boldsymbol{l}$ agrees with the network $G$  prediction $\bar{\boldsymbol{p}}$, where $\bar{\boldsymbol{p}}=argmax(\boldsymbol{p})$. If $\boldsymbol{p}_c$  exceeds the threshold $\tau$, the consistent set is defined as illustrated in Formula \ref{eq:1}:

\begin{equation}
	R = \mathbf{1} \big{[}\sum_{k=1}^K\left( \mathbf{1} \left[\bar{\boldsymbol{p}} =k \right]\cdot  \mathbf{1}\left[\boldsymbol{p}_k \geq \tau\right] \cdot \mathbf{1}\left[\boldsymbol{l}=k \right]\right)\big{]},
	\label{eq:1}
\end{equation}
where $\mathbf{1}$ is the indicator function, $K$ denotes the total number of categories, and $\tau$ is the confidence threshold. The generated $R$ is a binary mask.

The set of consistent point clouds $P^c$ and their respective pseudo-labels $\boldsymbol{l^c}$ are filtered by the mask $R$. Likewise, the ambiguous point cloud set $P^a$ and its associated pseudo-label $\boldsymbol{l^a}$ are determined by the complement of the mask $(1-R)$, as shown in Formulas \ref{eq:2} and \ref{eq:21}:
\begin{equation}
	P^c =R \cdot P,  \quad	P^a=(1-R) \cdot P,\\
	\label{eq:2}
\end{equation}
\begin{equation}
	\boldsymbol{l^c} = R \cdot \boldsymbol{l}, \quad \boldsymbol{l^a}=(1-R) \cdot \boldsymbol{l}.\\
	\label{eq:21}
\end{equation}

\subsection{Library of Prototypes}
To aid in learning prototype structures, we established two prototype memory banks, including a consistent prototype library and an ambiguous prototype library. The Exponential Moving Average (EMA) is employed to calculate the clustering centers when merging various batches, and then get the consistent and ambiguous prototypes driven by receiving multiple batches of samples. Initially, the consistent and ambiguous prototype memory banks are the clustering centers computed during clustering, i.e., $\boldsymbol{\mu}^c=\boldsymbol{\mu}^a=\boldsymbol{\mu}$. The specific update calculation for the prototype libraries are shown in Formulas \ref{eq:3} and  \ref{eq:31}:
\begin{equation}
	\boldsymbol{\mu}_k^c \leftarrow \alpha \boldsymbol{\mu}_k^c+(1-\alpha) \bar{\boldsymbol{\mu}_k^c},\\
	\label{eq:3}
\end{equation}
\begin{equation}
	\boldsymbol{\mu}_k^a \leftarrow \alpha \boldsymbol{\mu}_k^a+(1-\alpha) \bar{\boldsymbol{\mu}_k^a},
	\label{eq:31}
\end{equation}
where $\bar{\boldsymbol{\mu}_k^c}$ and $\bar{\boldsymbol{\mu}_k^a}$ represent the category $k$ clustering center feature representation in a single batch of samples. $\alpha$ is the EMA parameter, with a default setting of 0.99.

Enhancing the robust features in the prototype library involves calculating category centers for each batch and updating the library consistently. For a single batch, the consistent category clustering center $\bar{\boldsymbol{\mu}_k^c}$ and the ambiguous category clustering center $\bar{\boldsymbol{\mu}_k^a}$ can be defined as Formulas \ref{eq:4} and \ref{eq:41}:
\begin{equation}
	\bar{\boldsymbol{\mu}^c_k}^{(i)} =\frac{1}{\left|{\Omega_k^c}^{(i)}\right|} \sum_{\boldsymbol{p}_j^c \in {\Omega_k^c}^{(i)}} G(\boldsymbol{p}_j^c),\\
	\label{eq:4}
\end{equation}
\begin{equation}
	\bar{\boldsymbol{\mu}^a_k}^{(i)} =\frac{1}{\left|{\Omega_k^a}^{(i)}\right|} \sum_{\boldsymbol{p}_j^a \in {\Omega_k^a}^{(i)}} G(\boldsymbol{p}_j^a),\\ 
	\label{eq:41}
\end{equation}
where ${\Omega_k^c}^{(i)}$ and ${\Omega_k^a}^{(i)}$ are the set of points pseudo-labeled $k$ of $i$ batch for consistent and ambiguous points  respectively. This can be expressed as ${\Omega_k^c}^{(i)}=\{\boldsymbol{p}_j^c \mid l_j^c =k \}$ and ${\Omega_k^a}^{(i)}=\{\boldsymbol{p}_j^a \mid l_j^a =k \}$. Where $i=\left\{1, \ldots, B N_L\right\}$, $| \cdot |$ denotes the number of sets, and $B N_L$ denotes the number of batch.

\subsection{Consistent Structure Learning}
To leverage the potential of valid points, a structural error matrix is formulated by prototypes from a consistent prototype library with the consistent point feature of the Clustering centre for corresponding categories within a given batch. To learn consistent structures, the exact definition is depicted in Formula \ref{eq:5}:
\begin{equation}
	M_{k}^c =  ||\boldsymbol{\mu}_k^c-G(\boldsymbol{p}_j^c)||_2 ,
	\label{eq:5}
\end{equation}
where $M_{k}^c$ represents the structural error between the prototype of category $k$ in the consistent prototype library and the consistent feature within a single batch, with $||\cdot||_2$ denoting the Euclidean distance. 

Theoretically, minimizing the structural error facilitates the learning of consistent structures. Hence, the dependable structural loss can be represented by Formula \ref{eq:6}:
\begin{equation}
	\mathcal{L}_{s l}= \sum_{k=1}^K   M_{k}^c.
	\label{eq:6}
\end{equation}

Structure learning of consistent points is necessary, and the visualisation results in Fig. \ref{imgab2} validate our strategy. Consistent structure learning is equivalent to a reduction in the distance of a specific category from the prototype and a brightening of the colour.

\subsection{Semantic Relation Consistent Reasoning}
To constrain consistent and ambiguous points to have a consistent semantic representation. An artificial assumption is made that prototypical features acquired at consistent points are more precise than those obtained at ambiguous points.

Therefore, we introduce consistent and ambiguous prototypes from the consistent reasoning constraints prototype library. To maintain consistency in the similarity between consistent and ambiguous prototypes, distinct consistent and ambiguous similarity matrices are computed separately, as demonstrated in Formula \ref{eq:7}:
\begin{equation}
	e_{i j}^c = {\boldsymbol{\mu}_i^c}\cdot{\boldsymbol{\mu}_j^c}^T, \; e_{i j}^a = \boldsymbol{\mu}_i^a\cdot{\boldsymbol{\mu}_j^a}^T, \; i,j=\{1,2,\cdot\cdot\cdot,K\}.
	\label{eq:7}
\end{equation}

The similarity matrix can be normalized using Formula \ref{eq:8}:
\begin{equation}
	\bar{e}_{i j}^c=e_{i j}^c / \sum_{j=1}^K e_{i j}^c, \bar{e}_{i j}^a=log(e_{i j}^a / \sum_{j=1}^K e_{i j}^a).
	\label{eq:8}
\end{equation}

The relationship between consistent and ambiguous prototypes is often in error. For consistency preservation, it is possible to keep the information entropy between them relatively close. Hence, the consistent reasoning loss function is defined by Formula \ref{eq:9}:
\begin{equation}
	\mathcal{L}_{c r}=\frac{1}{K^2} \sum_{i=1}^{K} \sum_{j=1}^K  \bar{e}_{i j}^c  log (\bar{e}_{i j}^c/\bar{e}_{i j}^a). 
	\label{eq:9}
\end{equation}

Consistent reasoning extends the range of consistent points and enhances the robustness of the consistent prototype. In the visualisation results in Fig. \ref{imgab2}, as the reasoning continues, it turns out that the more ambiguous points that should have belonged to the category are classified as consistent points.

\subsection{Overall Objective Function}
In summary, the overall loss function can be formulated as Formula \ref{eq:14}:
\begin{equation}
	\mathcal{L}_{total} = \mathcal{L}_{ce} + \lambda_1\mathcal{L}_{sl} + \lambda_2\mathcal{L}_{cr},
	\label{eq:14}
\end{equation}
where $\lambda_1$ and $\lambda_2$ are hyperparameters.

\begin{table}
	\centering
	\resizebox{1.0\columnwidth}{!}{
		\begin{tabular}{c@{}rcccc}
			\toprule
			& Method & OA(\%) & mAcc(\%) & mIoU(\%) \\
			\midrule
			& PointNet & 77.5 & 59.1 & 44.6 \\
			Supervised	& PointNet++  & 77.5 & 62.6 & 50.1 \\
			Methods	& SparseConv  & 88.4 & 69.2 & 60.8 \\
			\midrule
			semi-supervised (10\%)	& Jiang et al.  & 69.1 & - & 57.7 \\
			\midrule
			weakly & MT  & - & - & 44.4 \\
			supervised (1pt) & Zhang et al. & - & - & 48.2 \\
			\midrule
			& KMeans & 22.1 & 21.2 & 9.4 \\
			& DBSCAN & 17.5 & 19.8 & 9.2 \\
			& IIC-S-PFH  & 31.2 & 16.3 & 9.1 \\
			Unsupervised& PICIE-S-PFH  & 48.4 & 40.4 & 25.2 \\
			Methods & GrowSP  & 78.4 & \textbf{57.2} & 44.5\\
			& U3DS$^3$  & 75.5 & 55.8 & 42.8\\
			& PointDC   & 54.1 & - & 22.6 \\
			& P-SLCR (ours) & \textbf{79.7} & \textbf{57.2} & \textbf{47.1}\\
			\bottomrule
	\end{tabular}}
	\caption{Quantitative results for semantic segmentation on the S3DIS dataset. Only 12 categories evaluated on the Area-5.}
	\label{tab:s3dis}
\end{table}

\subsection{Implementation Details}
The input point cloud generally comprises 6D raw features with normalized coordinates XYZ and color RGB. SemanticKITTI \cite{behley2019semantickitti} lacks color information. A straightforward and efficient neural network framework is implemented based on the SparseConv \cite{graham20183d} architecture, without any pre-training,  utilizing the KMeans algorithm for clustering hyperpoint features. Color enhancements \cite{liu2023cpcm} were applied, involving color shifting, contrast enhancement, and dithering techniques. The number of semantic primitives, $C$, must be defined for training purposes. $\lambda_1$ and $\lambda_2$ are 0 for the first half of the training phase and set to 1 for the second half. During testing, semantic categories of these primitives are clustered using the KMeans \cite{lloyd1982least} algorithm. Given the mismatch between the test labels and predicted classes, the  Hungarian algorithm aligns the predicted classes with the true labels.

\begin{figure}
	\centering
	\includegraphics[width=0.46\textwidth]{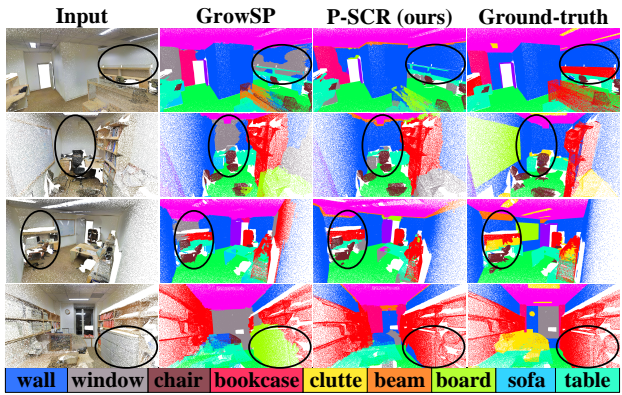} 
	\caption{The qualitative results on S3DIS demonstrate the segmentation performance in comparison to GrowSP. Each color represents a semantic class. We highlight the differences with black circles.}
	\label{imgvisS3dis}
\end{figure}
\begin{table}
	\centering
	\resizebox{0.9\columnwidth}{!}{
		\begin{tabular}{c@{}rcccc}
			\toprule
			& Method & OA(\%) & mAcc(\%) & mIoU(\%) \\
			\midrule
			Supervised & PointNet & 75.9 & 67.1 & 49.4 \\
			& PointNet++  & 77.1 & 74.1 & 55.1 \\
			Methods & SparseConv  &  89.4 & 78.1 & 69.2 \\
			\midrule
			& GrowSP  & 76.0  & 59.4 & 44.6  \\
			& P-SLCR (ours) &  \textbf{79.0} & \textbf{62.2} & \textbf{47.5}  \\
			\bottomrule
			
	\end{tabular}}
	\caption{Quantitative results for semantic segmentation on the S3DIS 6-Fold. Only 12 categories evaluated.}
	\label{tab:s3disTest}
\end{table}
\section{Experiments}
The effectiveness of our method was evaluated by testing it on three comprehensive point cloud datasets, encompassing both indoor \cite{armeni20163d, dai2017scannet} and outdoor \cite{behley2019semantickitti} scenarios. The overall accuracy (oAcc), the mean accuracy (mAcc), and the mean Intersection over Union (mIoU), which are commonly used in point cloud semantic segmentation methods, are mainly used as evaluation metrics. 
The comparison methods include PointNet \cite{qi2017pointnet}, PointNet++ \cite{qi2017pointnet++}, SparseConv \cite{graham20183d}, Zhang et al. \cite{zhang2021perturbed}, MT \cite{tarvainen2017mean}, Jiang et al. \cite{jiang2021guided}, KMeans \cite{lloyd1982least}, DBSCAN \cite{ester1996density}, IIC-S-PFH \cite{ji2019invariant}, PICIE-S-PFH \cite{cho2021picie}, PointDC \cite{chen2023pointdc}, GrowSP \cite{zhang2023growsp} and U3DS$^3$ \cite{liu2024u3ds3}.

\begin{figure}
	\centering
	\includegraphics[width=0.47\textwidth]{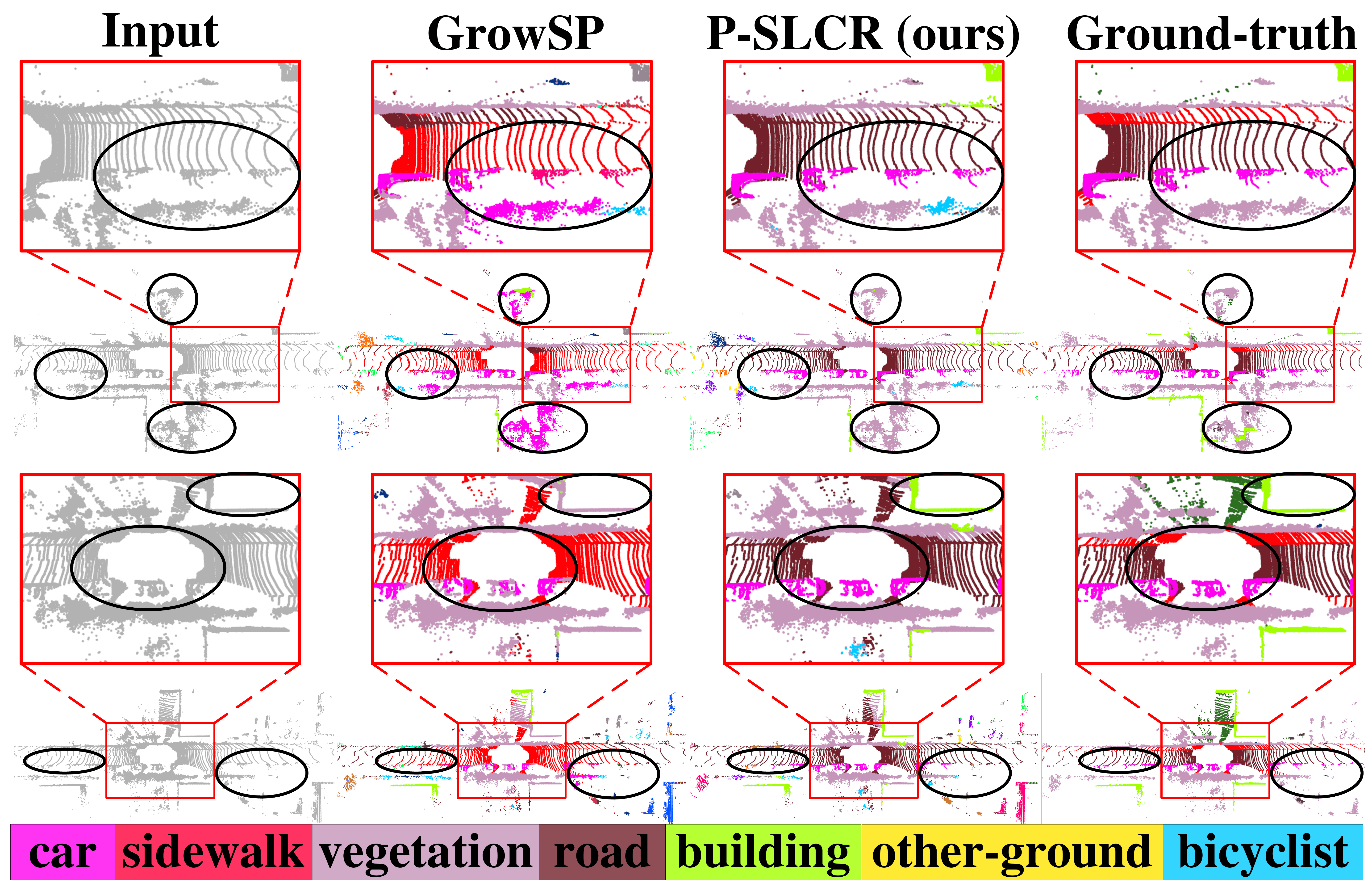} 
	\caption{The qualitative results on SemanticKITTI  demonstrate the segmentation performance in comparison to GrowSP. We zoomed in on local regions to show details and highlighted differences with black circles.}
	\label{imgvisKitti}
\end{figure}

\begin{table}
	\centering
	\resizebox{0.9\columnwidth}{!}{
		\begin{tabular}{c@{}rcccc}
			\toprule
			& Method & OA(\%) & mAcc(\%) & mIoU(\%) \\
			\midrule
			& KMeans  & 8.2 & 8.1 & 2.5 \\
			& DBSCAN  & 17.8 & 7.5 & 6.8 \\
			Unsupervised & IIC-S-PFH  & 23.4 & 9.0 &  4.6 \\
			Methods & PICIE-S-PFH  & 42.7 &  11.5 &  6.8 \\
			& GrowSP  & 38.3 & 19.7 & 13.2\\ 
			& U3DS$^3$ & 34.8 & \textbf{23.1} & 14.2\\
			& P-SLCR (ours) & \textbf{55.9} & 21.1 & \textbf{15.3}\\
			\bottomrule
			
	\end{tabular}}
	\caption{Quantitative results for semantic segmentation on the SemanticKITTI dataset. All 19 categories evaluated on the validation set.}
	\label{tab:kitti}
\end{table}

\begin{table}
	\centering
	\resizebox{0.7\columnwidth}{!}{
		\begin{tabular}{c@{}rccc}
			\toprule
			& Method & mIoU(\%) \\
			\midrule
			& PointNet  & 14.6 \\
			Supervised & PointNet++  & 20.1 \\
			Methods & SparseConv &  53.2 \\
			\midrule
			Unsupervised & GrowSP  &  14.3 \\
			Methods & P-SLCR (ours)   &  \textbf{15.9}  \\
			\bottomrule
			
	\end{tabular}}
	\caption{Quantitative results for semantic segmentation on the SemanticKITTI dataset. All 19 categories evaluated on the online test set.}
	\label{tab:kittiTest}
\end{table}

\subsection{Evaluation on S3DIS}
S3DIS \cite{armeni20163d} is a large-scale indoor point cloud dataset divided into 6 areas, labeling the point cloud into 13 classes with one clutter class. For comparison with previous methods, we remove clutter classes during testing. It is also compared with fully supervised methods, semi-supervised methods, weakly supervised methods and the latest unsupervised methods.

The results for S3DIS Area-5 are shown in Table \ref{tab:s3dis}, our method (P-SLCR) achieves the best performance for all metrics in the unsupervised method comparison. Compared to the next best unsupervised method, GrowSP, P-SLCR  outperforms with a 1.3\% higher OA and a 2.6\% higher mIoU. P-SLCR remains competitive even when compared with weakly supervised methods. In addition, it is worth noting that P-SLCR outperforms the classical fully supervised network PointNet for the first time and by 2.5\% in mIoU. The results for S3DIS 6-Fold are shown in Table \ref{tab:s3disTest}, P-SLCR is substantially ahead of GrowSP in all experimental metrics.

Fig. \ref{imgvisS3dis} presents qualitative results obtained by the state-of-the-art unsupervised method GrowSP and our proposed P-SLCR. GrowSP tends to misclassify walls and bookcases as windows or boards, whereas P-SLCR produces more accurate and semantically consistent segmentation results.



\begin{figure}
	\centering
	\includegraphics[width=0.45\textwidth]{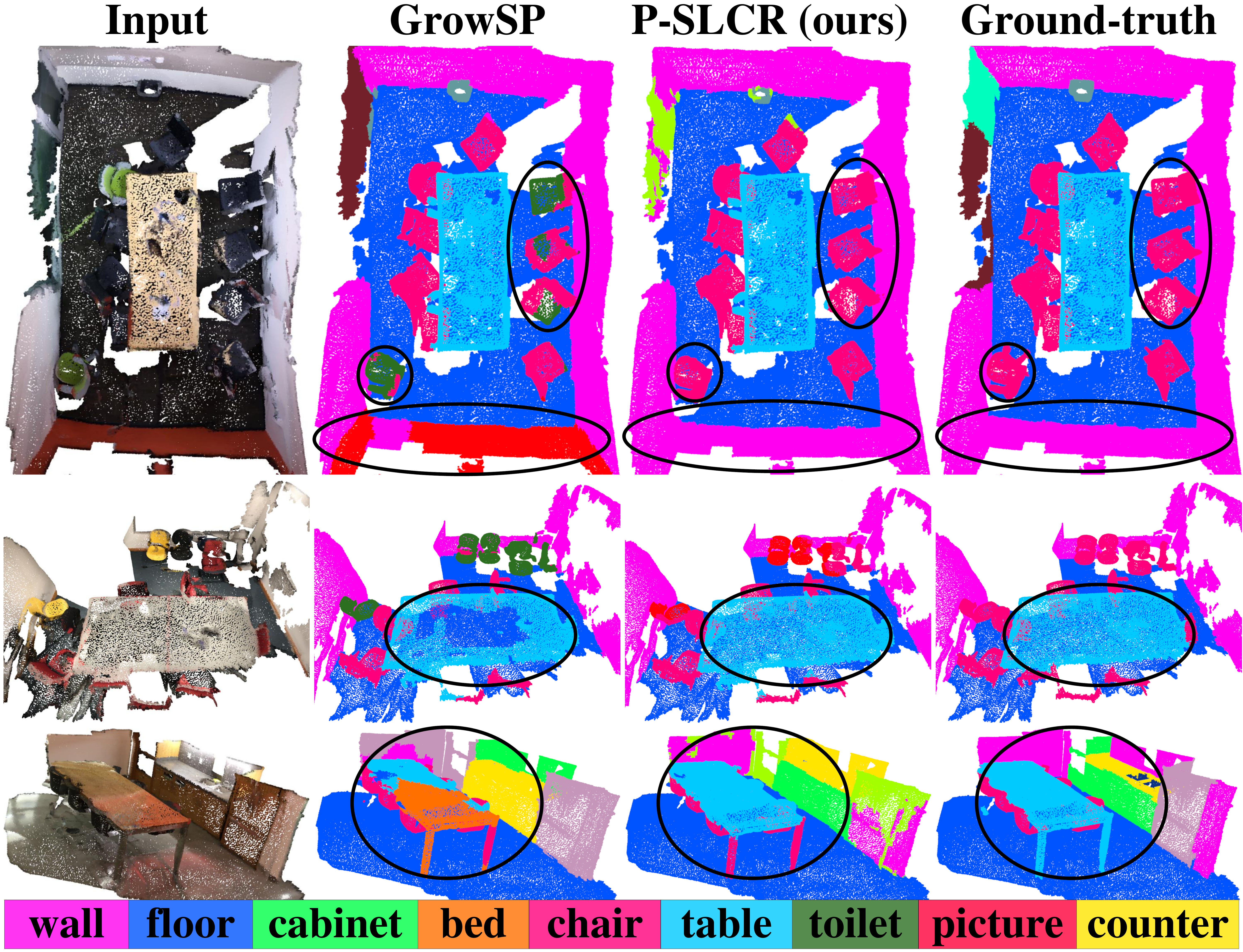} 
	\caption{The qualitative results on ScanNet demonstrate the segmentation performance in comparison to GrowSP. Each color represents a semantic class. We highlight the differences with black circles.}
	\label{imgvisScannet}
\end{figure}
\subsection{Evaluation on SemanticKITTI}
SemanticKITTI \cite{behley2019semantickitti} is a large outdoor driving dataset collected using laser LIDAR. In total, there are 22 sequences and 19 semantic categories. Comparisons are made primarily with unsupervised methods that have migrated from other fields, as well as with more recent unsupervised point cloud methods.

As shown in Table \ref{tab:kitti}, our method (P-SLCR) obtains the optimal OA and mIoU. It is 0.9\% ahead of the U3DS$^3$ \cite{liu2024u3ds3}, in terms of mIoU. Notably, it is substantially ahead of other unsupervised methods in OA, about 20\% ahead of the state-of-the-art unsupervised point cloud methods.  The results for SemanticKITTI online test are shown in Table \ref{tab:kittiTest}, P-SLCR outperforms GrowSP by 1.6\% on mIoU.

Fig. \ref{imgvisKitti} presents the visual outcomes obtained through the latest unsupervised method, GrowSP \cite{zhang2023growsp}. GrowSP tends to misclassify extensive road  segments as sidewalk, a discrepancy effectively addressed by P-SLCR.  Moreover, GrowSP struggles with distinguishing between cars and vegetation, leading to semantic confusion. It frequently separates cars and vegetation erroneously, a challenge absent in P-SLCR, underscoring the robustness of P-SLCR.

\begin{table}
	\centering
	\resizebox{1.0\columnwidth}{!}{
		\begin{tabular}{c@{}rcccc}
			\toprule
			& Method & OA(\%) & mAcc(\%) & mIoU(\%) \\
			\midrule
			& KMeans  & 10.2 & 10.4 & 3.4 \\
			& DBSCAN  & 15.3 & 10.1 & 6.1 \\
			Unsupervised& IIC-S-PFH  & 18.9 &  6.3 &  3.0 \\
			Methods & PICIE-S-PFH  & 23.6 &  15.1 &  7.4 \\
			& GrowSP  & 57.3 & 44.2 & 25.4 &\\ 
			& U3DS$^3$  & 60.1 & 46.8 & 27.3\\
			& P-SLCR (ours) & \textbf{61.4} & \textbf{49.0} & \textbf{29.0}\\
			\bottomrule
			
	\end{tabular}}
	\caption{Quantitative results for semantic segmentation on the ScanNet dataset. All 20 categories evaluated on the validation set.}
	\label{tab:scannet}
\end{table}

\begin{table}
	\centering
	\resizebox{1.0\columnwidth}{!}{
		\begin{tabular}{c@{}lcccc}
			\toprule
			&  & OA(\%) & mAcc(\%) & mIoU(\%) \\
			\midrule
			(1) \; & Remove Reliable Structure Learning & 79.36 & 55.63 & 45.65 \\
			(2) \; & Remove Consistent Reasoning & 79.08 & 54.09 & 42.37 \\
			(3) \; & Remove Color Enhancement & 79.28 & 57.50 & 45.21 & \\
			(4) \; & $\tau $ = 0.6 for Degree of Reliability & 79.19 & 57.08 & 43.80\\
			(5) \; & $\tau $ = 0.7 for Degree of Reliability & 79.72  & 57.23 & \textbf{47.08} \\
			(6) \; & $\tau $ = 0.8 for Degree of Reliability &  78.88 & 56.76 & 44.50 \\
			(7) \; & Number of Semantic Primitive = 200  & 78.81 & \textbf{58.45} & 44.25 \\
			(8) \; & Number of Semantic Primitive = 300 & 79.72  & 57.23 & \textbf{47.08} \\
			(9) \; & Number of Semantic Primitive = 400 & 78.01 & 58.36 & 45.27 \\
			
			(10) & $\mathcal{L}_{cr}$ Change to using Cross-Entropy & \textbf{80.27} & 56.63 &  44.72 \\ 
			
			\bottomrule
	\end{tabular}}
	\caption{The OA, mAcc and mIoU scores of all ablated networks on Area-5 of S3DIS based on our full P-SLCR.}
	\label{tab:newab}
\end{table}

\subsection{Evaluation on ScanNet}
ScanNet \cite{dai2017scannet} is an indoor dataset in RGB-D captured from real scenes, with 41 categories in the training set and 20 categories in the validation and test sets. The main comparison is with state-of-the-art unsupervised point cloud semantic segmentation methods.

As shown in Table \ref{tab:scannet}, our method (P-SLCR) achieved the best performance in all metrics, with 1.3\%, 2.2\%, and 1.7\% improvement in OA, mAcc, and mIoU, respectively, over the next best method  \cite{liu2024u3ds3}.

Fig. \ref{imgvisScannet} displays that P-SLCR accurately segments the categories of chairs, tables, and walls. GrowSP often incorrectly divides objects, especially smaller categories like chairs and tables, resulting in a single class being split into multiple categories.


\subsection{Ablation Study}
A series of ablation studies were conducted on the S3DIS \cite{armeni20163d} dataset Area-5  to assess the efficacy of our proposed components and hyperparameter selection.Four groups of ablation studies were carried out, as illustrated in Table \ref{tab:newab}.  


\begin{figure}
	\centering
	\includegraphics[width=0.47\textwidth]{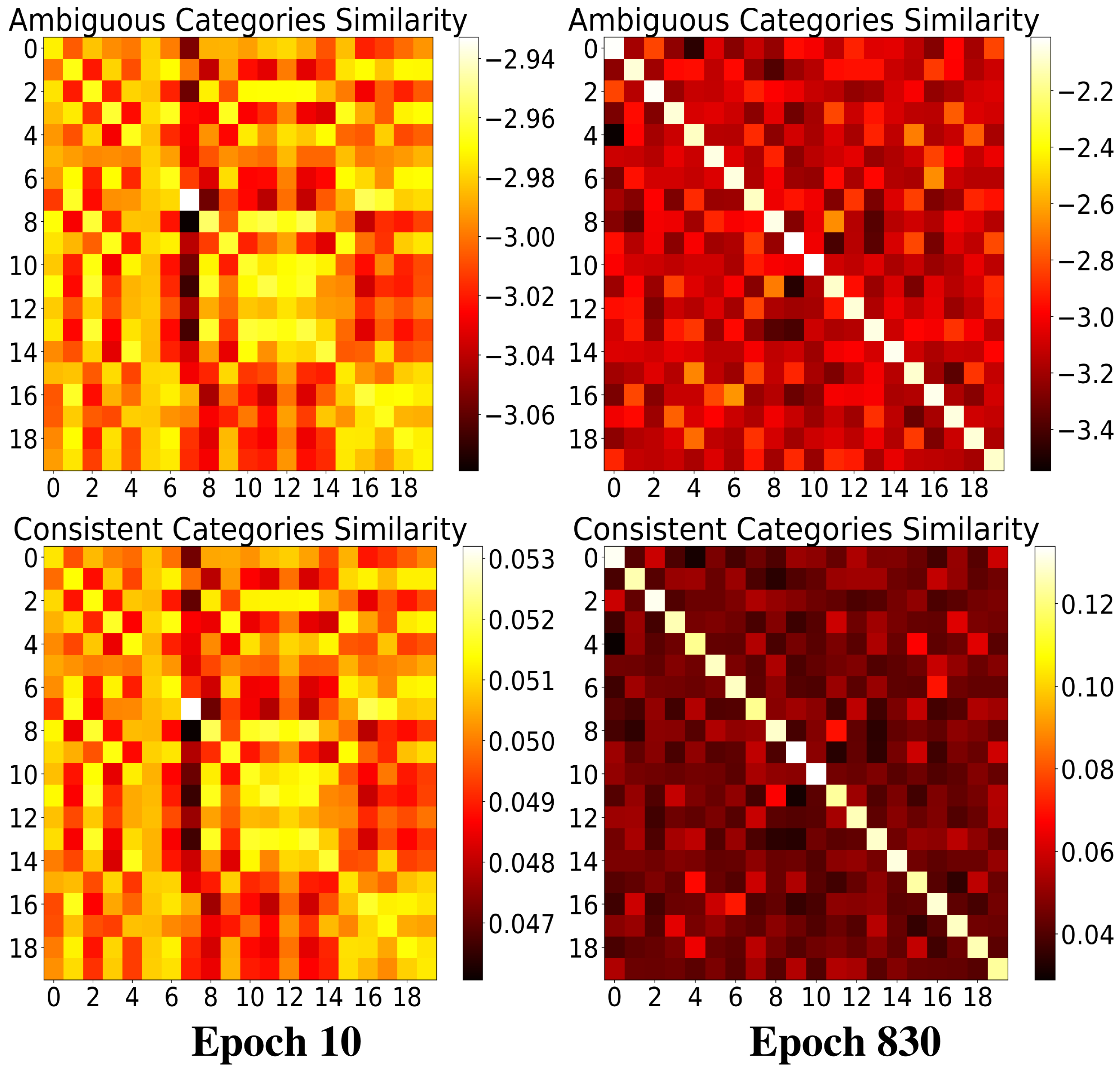} 
	\caption{Similarity measure matrix visualization results for consistent and ambiguous prototypes. For presentation purposes, we clustered the number of prototypes into categories of visualization results for ScanNet validation sets.}
	\label{imgab1}
\end{figure}


\textbf{(1)-(3) Contribution of Components.} The contributions of each component to the complete method are illustrated by individually removing them. The degree of reliability and semantic primitive category is set to 0.7 and 300, respectively, and other settings are consistent with the full method. Removing consistent reasoning has the greatest impact on the model, with mIoU plummeting by 4.71\%.

\textbf{(4)-(6) Degree of Reliability.} The selection of dependable points is influenced by the sensitivity to different levels of reliability. Typically, with increasing experience, the reliability threshold is set above 0.5. In this study,  we have considered three specific values: 0.6, 0.7, and 0.8. A notable advantage is observed at a threshold of 0.7, leading us to adopt this value as the degree of reliability in our comprehensive methodology.

\textbf{(7)-(9) Number of Semantic Primitive.} We investigated the sensitivity to the number of distinct semantic primitive categories, as indicated by the number of clustering centers. Within this framework, we examined three values for comparison: 200, 300, and 400. Notably, a substantial advantage was observed at 300, leading to the adoption of this value as the number of semantic primitive categories in our complete.

\textbf{(10) $\mathcal{L}_{cr}$ Change to using Cross-Entropy.} Replacing the constraint with cross-entropy loss caused a sharp drop in mIoU. It shows that excessive constraints on the similarity distribution between consistent and ambiguous and prototypes are detrimental to semantic recognition.

\begin{figure}
	\centering
	\includegraphics[width=0.45\textwidth]{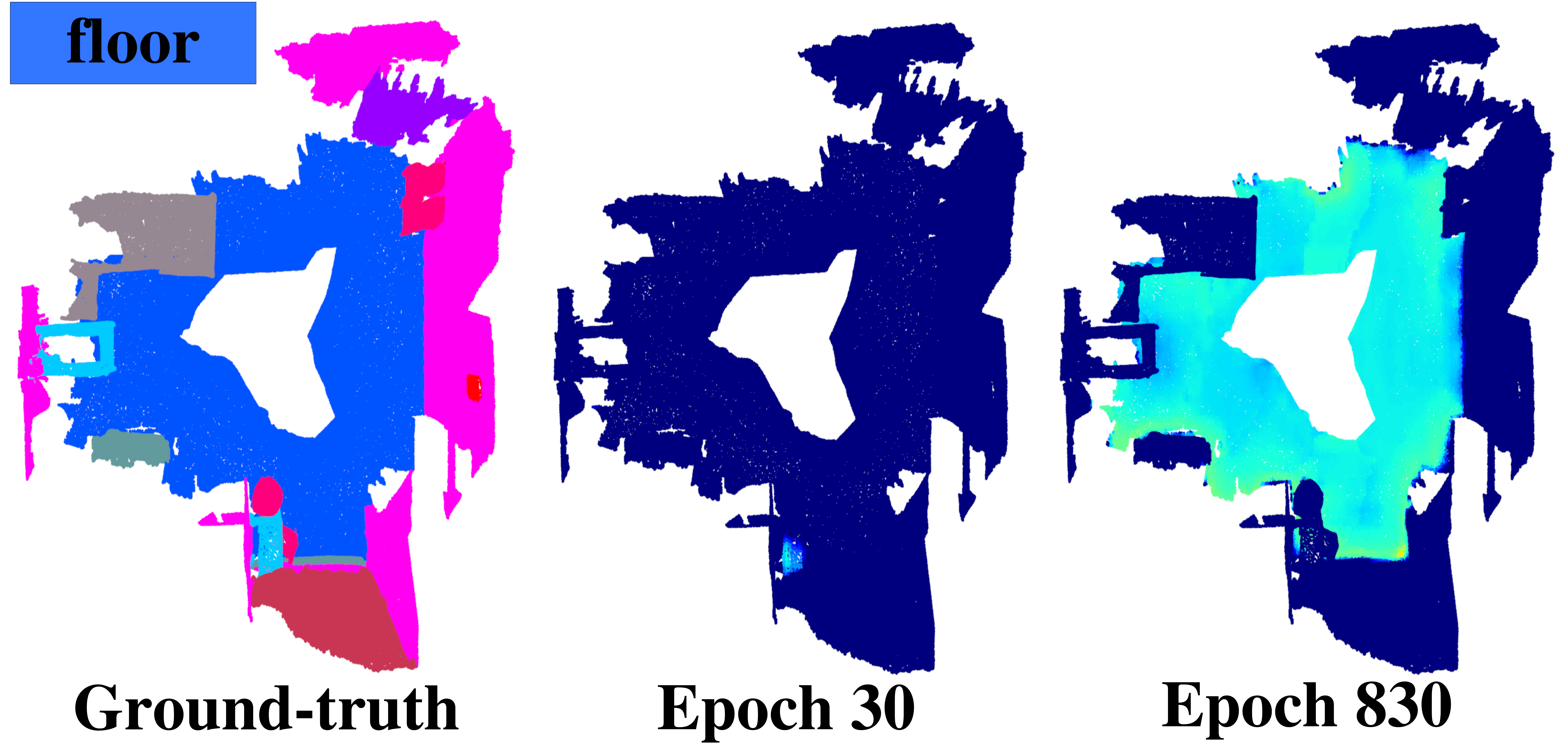} 
	\caption{The visualization shows the distance from the prototype for different categories of consistent points. The visualization results for the validation set based on the ScanNet.}
	\label{imgab2}
\end{figure}

		%

\textbf{Analysis}:
Fig. \ref{imgab1} illustrates the similarity measure matrices of both consistent and ambiguous prototypes. At epoch 10, the similarity range is minimal, suggesting little distinction among prototype categories.  Introducing the P-SLCR method led to heightened discrepancies among prototypes by epoch 830, revealing prominent features. Notably,  the similarity to the prototype itself significantly surpassed the similarity to other categories. Comparison with the unsupervised point cloud semantic segmentation methods demonstrates the state-of-the-art of our approach (P-SLCR). However, we found that the accuracy of unsupervised methods is still low, especially when there is no color information. During training, the model seems to increasingly prioritize approaching areas with higher point density, potentially leading to the neglect of certain correct edges (Fig. \ref{imgab2}). (See the Appendix for more discussion)

\section{Conclusion}
We propose a new architecture for prototypes consistent structure learning and consistent reasoning model (P-SLCR), including a library of learnable prototypes. The pseudo-labels generated during clustering, the training phase selects consistent points by reliability for structure learning and constructs semantic consistent reasoning between consistent and ambiguous prototypes. We validated P-SLCR against the latest models using the S3DIS, ScanNet, and SemanticKITTI datasets. The results confirm that our unsupervised approach surpasses the fully supervised method, introducing a novel framework for 3D unsupervised learning. Our future work will continue to explore the potential of P-SLCR for other 3D unsupervised tasks.

\bibliography{aaai2026}

\end{document}